\documentclass[10pt,twocolumn,letterpaper]{article}

\usepackage{iccv}
\usepackage{times}
\usepackage{epsfig}
\usepackage{graphicx}
\usepackage{amsmath}
\usepackage{amsthm}
\usepackage{float}
\usepackage{amssymb}
\usepackage[ruled,vlined,noend]{algorithm2e}

\usepackage[pagebackref=true,breaklinks=true,letterpaper=true,colorlinks,bookmarks=false]{hyperref}

 \iccvfinalcopy 

\def\iccvPaperID{***} 

\ificcvfinal\fi

\begin{document}

\title{ PAL : Pretext-based Active Learning}
\author {
     \textbf{Shubhang Bhatnagar},\textsuperscript{\rm 1} \textbf{Sachin Goyal},\textsuperscript{\rm 2} \textbf{Darshan Tank},\textsuperscript{\rm 1}  \textbf{Amit Sethi} \textsuperscript{\rm 1} \\
     \textsuperscript{\rm 1}Indian Institute of Technology, Bombay \\
     \textsuperscript{\rm 2}Microsoft Research, India \\
 }

\maketitle
\ificcvfinal\thispagestyle{empty}\fi
\begin{abstract}


The goal of pool-based active learning is to judiciously select a fixed-sized subset of unlabeled samples from a pool to query an oracle for their labels, in order to maximize the accuracy of a supervised learner. However, the unsaid requirement that the oracle should always assign correct labels is unreasonable for most situations. We propose an active learning technique for deep neural networks that is more robust to mislabeling than the previously proposed techniques. Previous techniques rely on the task network itself to estimate the novelty of the unlabeled samples, but learning the task (generalization) and selecting samples (out-of-distribution detection) can be conflicting goals. We use a separate network to score the unlabeled samples for selection. The scoring network relies on self-supervision for modeling the distribution of the labeled samples to reduce the dependency on potentially noisy labels. To counter the paucity of data, we also deploy another head on the scoring network for regularization via multi-task learning and use an unusual self-balancing hybrid scoring function. Furthermore, we divide each query into sub-queries before labeling to ensure that the query has diverse samples. In addition to having a higher tolerance to mislabeling of samples by the oracle, the resultant technique also produces competitive accuracy in the absence of label noise. The technique also handles the introduction of new classes on-the-fly well by temporarily increasing the sampling rate of these classes.

\end{abstract}
\section{Introduction}

In spite of their unprecedented accuracy on several tasks involving image analysis, a hurdle in using deep convolutional neural networks (CNNs) for many real problems is their requirement of large labeled datasets. Labeling and annotations are laborious and costly for several domains, such as medical imaging, where follow-up or expertise is required. Strategies to reduce the number of labels include transfer, semi-supervised, weakly-supervised, few-shot, and active learning. Active learning algorithms are used to decide whether or not to send an unlabeled sample for labeling to an oracle (e.g., a radiologist for x-ray images), such that the increase in the task performance (e.g., classification accuracy) is maximized with respect to a labeling cost. Active learning can be used select the best data for labeling even in conjunction with the some of the other techniques.

In pool-based active learning, training progresses iteratively in rounds starting from an unlabeled pool of samples. In each round, up to a budgeted number of $N$ additional samples can be selected from the unlabeled pool for labeling \cite{ducoffe,Sinha_2019_ICCV,sener2018active}. After a random initial selection of samples for labeling, the selection strategy is usually based on picking \emph{novel} and \emph{diverse} samples from the unlabeled pool. Novelty (a.k.a. \emph{uncertainty} and \emph{confusion}) refers to selecting samples that are least similar to the previously labeled samples in order to maximize the information gain by getting them labeled. Diversity refers to selecting samples that maximize the collective information gained by the labeling of the samples in a query, instead of simply selecting novel samples that may be similar to each other.

A majority of the previously proposed active learning methods have relied on the task network itself for estimating uncertainty \cite{Sinha_2019_ICCV,gal2017deep,ducoffe,Beluch_2018_CVPR}. However, it recently became clear that a task network is a poor estimator of its own uncertainty on unlabeled samples that are unlike the labeled samples \cite{hein2019relu}. We, therefore, use a second network -- called the \emph{scoring network} -- just for the purpose of scoring unlabeled samples for their novelty and diversity (Section~\ref{sec:method}), in line with a few previous studies \cite{yoo2019learning}. 

Secondly, modeling the distribution of the labeled data to identify novel (out-of-distribution) unlabeled samples relies on the labels gathered so far. However, it is quite realistic to assume that the oracle mislabels a certain fraction of samples sent to it due to factors such as human subjectivity and error. This exacerbates the problem of relying on the task network to pick unlabeled samples for the query. Although we also model the distribution of the labeled data, our main contribution is to do so without relying on the potentially noisy labels too much. We train the scoring network on the labeled samples for a self-supervised learning (SSL) task (Section~\ref{sec:SSL}), specifically, to predict random rotations of an image. We use the extent to which an unlabeled sample gives a wrong prediction on the SSL task as an indicator of its novelty. The self-supervision labels can be generated inexpensively for testing the uncertainty of the unlabeled samples, compared to several other techniques. Due to the use of the pretext (SSL) task, we call this scheme \emph{pretext-based active learning (PAL)}.

Thirdly, SSL works best with a large dataset but the labeled pool in active learning is usually small. Therefore, we train the scoring network with another head for multi-task learning, which is known to regularize neural networks when the training data for each task is limited~\cite{ben2008notion}. An additional advantage of a multi-task scoring network is that the mistakes of the SSL head can be compensated for by the task head. For example, if an image has rotational symmetry the rotation prediction SSL task might fail, but the image may still be easily classifiable. Furthermore, our formulation of the scoring function allows for a self-adjusting trade-off between the two heads for mutual correction (Section~\ref{sec:hybrid}).

Finally, we ensure diversity among the $N$ samples selected during a query by breaking it into $K$ sub-queries. For each sub-query, we pick samples that are novel with respect to the previous sub-queries, which ensures diversity among the samples of the query itself. We tune only the self-supervision head between the sub-queries, so that we do not incur the labeling cost until the entire query is formulated (Section~\ref{sec:diversity}).

Due to the design features mentioned above, PAL showed an accuracy that is competitive with the state-of-the-art \cite{Sinha_2019_ICCV,sener2018active,gal2017deep} on benchmark image recognition and segmentation datasets, without using a computationally expensive training scheme (Section \ref{sec:main_res}). More importantly, PAL seems to be significantly more robust to partial mislabeling of the training data by the oracle (Section~\ref{sec:robust}). 

We also tested PAL for a scenario called \emph{biased initial pool}, in which certain classes may be underrepresented (or absent) in the initially labeled data. As desired, PAL over-samples the previously underrepresented classes and ramps up the performance on them in the first few queries itself, and then returns to balanced sampling (Section~\ref{sec:bias_pool}).

We also show that PAL is insensitive to the choice of the scoring network architecture (Section~\ref{sec:Robust2Arch}), but each of its three distinct components -- self-supervision, supervision, and diversity -- are individually important (Section~\ref{sec:ablation}).

\section{Related Work}

\subsection{Active learning}

There are several settings for active learning, such as \emph{membership query synthesis} and \emph{stream-based sampling}. In the former, the learner generates new samples to query the oracle \cite{anguin, gaal, huijser}, while in the latter the unlabeled dataset is presented as a stream, and is evaluated online \cite{cohn, dasgupta}. However, unlike these settings, the proposed method is \emph{pool-based sampling}, which makes a complete use of labeled and unlabeled data pools, when the latter is also available~\cite{Sinha_2019_ICCV,sener2018active,gal2017deep}. In this setting, starting with a set of labeled samples, unlabeled samples of a budgeted number are selected for querying the oracle for their labels.

Pool-based active learning techniques aim to pick samples that are novel and diverse. \textbf{Novelty} (a.k.a., uncertainty, confusion, perplexity, non-triviality, out-of-distribution, and informativeness) refers to an unlabeled sample's ability to provide new information, if labeled, \emph{independently of other samples selected}. Some of the early measures of novelty have known issues. For example, entropy of the estimated class probability mass function~\cite{settles} is prone to calibration error~\cite{hein2019relu}, discordance between a committee of classifiers~\cite{qbc} can be computationally expensive, and distance from a linear decision boundary~\cite{tong2001support} is not directly applicable to CNNs because of their complex decision boundaries. Distance from an adversarial example has been proposed as an approximation of distance from decision boundary \cite{ducoffe}, but also it requires computationally expensive gradient descent on image pixels. Uncertainty estimations based on Bayesian frameworks, such as MC-dropout~\cite{bay1, bay2}, are also computationally expensive. Surprisingly, no one has used the difficulty of solving a self-supervised (pretext) task as a measure of novelty, which requires only up to one additional network to be trained in parallel with the task network.

Methods based on \textbf{diversity} (a.k.a. representativeness and coverage) seek to select samples that can represent the unlabeled data distribution well. If the samples selected in a query are individually novel with respect to the previously labeled samples but collectively similar to each other, then the joint information gained from their labels as a query group may not be maximized. A method based on identifying a \emph{core-set} has been proposed that models the empirical loss over the set of already labeled samples combined with the pool of query samples on the empirical loss over the whole dataset \cite{sener2018active}. However, this approach suffers when the representations are high-dimensional, because the Euclidean distance is a poor local similarity estimator in high dimensional spaces. An alternative approach called variational adversarial active learning (\emph{VAAL}) aims to learn a good representation using a variational autoencoder (VAE) trained adversarially using a discriminator that tries to predict if a sample is already labeled \cite{Sinha_2019_ICCV}. However, this is also a computationally expensive technique due to VAE training.

\subsection{Self-supervised learning}

Self-supervised learning (SSL) has shown great promise in learning usable data representations without needing explicit data labels. The learned representation can later be fine-tuned with a smaller labeled dataset. Many of the proposed SSL techniques automatically create a supervised pretext task by degrading an unlabeled image, and training a neural network to recover the original image. Some commonly used randomized degradations on an image for SSL are removing color~\cite{color}, reducing resolution~\cite{superres}, occluding parts of an image~\cite{inpainting}, jumbling the spatial order of its sub-images~\cite{jigsaw}, and applying  random geometric transforms~\cite{gidaris2018unsupervised}. Several other recent SSL techniques are based on contrastive learning, like SimCLR\cite{chen2020simple} and MoCo\cite{he2020momentum} instead focus on making a CNN learn image representations that are closer for augmented versions of the same images compared to those of the others.

If a CNN trained using an SSL task can correctly solve the SSL puzzle on a test image, it can be interpreted that the test image is similar to some of the training images~\cite{hendrycks2019using, golan}. Since training the scoring network using SSL on labeled samples does not require the oracle's labels, which may be noisy, we use the difficulty of solving the SSL task as a robust measure of the novelty of unlabeled images. However, SSL requires a large dataset, but in active learning the set of labeled images is small.  Additionally, sometimes the SSL task cannot be solved even for non-novel images, such as prediction the synthetic rotation of a rotationally-symmetric image.

\subsection{Multi-task learning}

It has been shown that if tasks are related and training dataset is limited, multi-task learning can help improve the accuracy of both tasks~\cite{ben2008notion}. Consequently, multi-task learning has been a subject of vigorous research in deep learning~\cite{zhang2014facial,bingel2017identifying,seltzer2013multi,doersch2017multitask}. In our experience, multi-task learning often requires smaller computational costs in comparison to ensembles for comparable gains in accuracy. We use multi-task learning in the scoring network by adding another head for classification. This head potentially also compensates for the mistakes of the SSL head on non-novel images that are difficult for SSL.

\section{Method: Pretext-based Active Learning}\label{sec:method}

Our method is an instance of pool-based active learning, essence of which can be described as follows. Let the pool of the currently labeled samples be $\mathcal{D}_{L}$ and the pool of unlabeled samples be $\mathcal{D}_{U}$. A task network $f_{\theta}(\textbf{x}_l)$ parameterized by $\theta$ is trained on all samples $\textbf{x}_l \in \mathcal{D}_{L}$. The active learning algorithm selects a budgeted set of $N$ or fewer samples from $\mathcal{D}_{U}$ in each query. The queried samples are then labeled by an oracle (assumed ideal, although unrealistic), added to $\mathcal{D}_{L}$, and removed from $\mathcal{D}_{U}$. The task network is retrained on the expanded $\mathcal{D}_{L}$ and its increase in accuracy is examined. This process is repeated until a specified number of samples $|\mathcal{D}_{L}|$ are labeled or a desired accuracy level is achieved. 

Previous methods estimated the uncertainty of the unlabeled samples using the task network itself, e.g., based on the entropy of computed class probabilities. We use a different neural network than task network for our selection strategy, which we refer to as the \emph{scoring network} hereafter. The scoring network has two heads, one for self-supervision and another for classification, whose outputs are used to assign a confusion score $S$ to an unlabeled image $\textbf{x}_u$.

\subsection{Self-supervision head}\label{sec:SSL}

The self-supervision head estimates the likelihood of the unlabeled data under the distribution of the labeled samples. We estimate this likelihood using the self-supervision score $S_{S}$, which is based on the prediction of randomized rotation for classification (SSL)\cite{gidaris2018unsupervised}, and SimCLR based on contrastive learning~\cite{chen2020simple} for semantic segmentation.

In the rotation task, we rotate the images by $90i^{\circ}$ for $i \in \{0,1,2,3\}$ and train a network $g_{\phi}$ parameterized by $\phi$ to predict $i$ on only the images from $\mathcal{D}_{L}$, so that the head learns the distribution of the labeled data. Using the self-supervised head, the following confusion score $S_S$ is assigned to each unlabeled image $\textbf{x}_u$: 
\begin{equation} S_S(\textbf{x}_u)=-\sum_{i \epsilon \{0,1,2,3\}} g_{\phi}(\text{rot}_{90i}(\textbf{x}_u))_{i}, 
\label{eq:SSScore}
\end{equation}
where $\text{rot}_{90i}(.)$ is the rotation function and $g_{\phi}(.)_{i}$ is the estimated probability of the $i^{\text{th}}$ rotation angle. We hypothesize that an image $\textbf{x}_{u} \in \mathcal{D}_{U}$ for which $S_S$ is closer to its minimum possible value of $-4$ will likely be similar to the labeled points in $\mathcal{D}_{L}$, and will fetch little extra information, if labeled. Conversely, for novel points $S_S$ will be closer to $0$.

In SimCLR~\cite{chen2020simple}, we apply random transformations including crops, resizes, and color jitter to the image, and train the network $g_{\phi}$ to reduce the distance between embeddings of two such transformed versions of the image. We train the network $g_{\phi}$ in this manner only on images belonging to $\mathcal{D}_{L}$, making the network learn the distribution of the labeled data. Specifically, for the semantic segmentation task, we use the embedding obtained from the bottleneck layer of the segmentation architecture to perform our contrastive learning task. Using the self-supervised head, we define the confusion score $S_{S}$, assigned to each unlabeled image $\textbf{x}_u$ as:
\begin{equation} S_S(\textbf{x}_u)=sim(g_{\phi}(\textbf{x}_{u,1}),g_{\phi}(\textbf{x}_{u,2}))
\label{eq:SSScore2}
\end{equation}
where $\textbf{x}_{u,1}$ and $\textbf{x}_{u,2}$ are transformed versions of $\textbf{x}_{u}$, and $sim(a,b)$ denotes the cosine similarity between 2 vectors $a$ and $b$. An image $\textbf{x}_{u} \in \mathcal{D}_{U}$ which has $S_{S}$ closer to its minimum of 0 will most likely be similar to points in $\mathcal{D}_{L}$, and will fetch less extra information. We think that other self-supervised tasks can also be used to generate a suitable estimate $S_{S}$.

\subsection{Classification head and hybrid score}\label{sec:hybrid}

A scoring network trained just on the self-supervised task may not work very well for the following reasons. Firstly, SSL requires a large number of training samples and should be trained from scratch. Moreover, the score $S_S$ is not reliable for images that have a rotational symmetry, and might be high even if the scoring network has modeled the semantic features well. Furthermore, the labels of $\mathcal{D}_L$, that represent at least partially reliable information, are left unused by the scoring network. To counter these issues, we introduce a classification head $h_{\psi}(\textbf{x}_u)$ parameterized by $\psi$ in the scoring network. We compute the degree to which the outputs of $h_{\psi}$ for an unlabeled sample $\textbf{x}_u$ are close to a uniform distribution $U$, using KL divergence (or relative entropy)~\cite{hendrycks2016baseline}, as a second measure of confusion $S_{C}(\textbf{x}_u)$, to give a \emph{hybrid confusion score} $S(\textbf{x}_{u})$, as shown below:
\begin{align}\label{eq:score}
 & S(\textbf{x}_{u})=S_S(\textbf{x}_{u}) +  \lambda S_C(\textbf{x}_{u}) \text{,  where} \\
 & S_C(\textbf{x}_{u})=- \text{KL}(U\ ||\ h_{\psi}(\textbf{x}_{u})),
\nonumber
\end{align}
where $\lambda \geq 0$ is a relative importance hyperparameter. When applying PAL to semantic segmentation, we calculate $S_{C}$ pixel-wise and average it to get $S_{C}$ for the sample. For a novel sample, we expect the KL divergence to be low and $S_C$ to be high.

Although, the entropy of class probabilities is a more popular measure of confusion~\cite{settles}, its range is finite. Had it been used as $S_C$ in place of the negative of KL divergence in Equation~\ref{eq:score} it would not have been able to counter-balance the effect of $S_S$ (Equation~\ref{eq:SSScore}) when it fails (e.g., in case of rotational symmetry). On the other hand, when the class prediction by $h_{\psi}$ is very confident and KL-divergence is high, that means, as desired,we would rely less on the SSL task, and combined score will self-adjust due to the infinite range of the KL-divergence.  

An added advantage of using a multi-task setting for the scoring network is getting better ordinal estimates of a true latent score due to an ensemble-like effect. This reasoning is valid as long as the correlations between the two components of the score and their correlation with the underlying score are positive. The former is empirically true, as shown in Table~\ref{tab:corr}, and the latter is likely true due to the training. We select the $N$ most informative samples from $\mathcal{D}_U$ with the highest $S(\textbf{x}_{u})$ as per Equation~\ref{eq:score} in each query round, after finding a good setting for the hyperparameter $\lambda \geq 0$ based on validation.


\begin{table}[]
    \centering
    \caption{Pearson's correlation ($r_{p}$) and Spearman's correlation ($r_{s}$) between $S_S$ and $S_C$ of unlabeled data points for the four datasets using a model trained on 10\% of the samples.}
    \label{tab:corr}
    \vspace{0.1cm}
    \begin{tabular}{|c|c|c|c|c|}
    \hline
         \ \  & SVHN & CIFAR-10 & Caltech-101  \\ \hline
          $r_{p}$ & 0.42 & 0.49 & 0.63 \\ \hline
         $r_{s}$ & 0.44 & 0.50 & 0.57 \\ \hline
    \end{tabular}
\end{table}

\subsection{Diversity score}\label{sec:diversity}

To ensure that the $N$ samples in a query are diverse to cover the unlabeled data distribution, we divide the query into $K$ sub-queries with $\frac{N}{K}$ samples each. For selecting the first sub-query, we select the top $\frac{N}{K}$ samples using the confusion score $S$ from Equation~\ref{eq:score}. After picking these, we fine-tune the scoring network using self-supervision without asking the oracle for their labels in the middle of the query. Using this fine-tuned network, we generate a score $S_{D}$, quite similar to Equation~\ref{eq:SSScore}, where $g_{\phi'}$ replaces $g_\phi$. $S_{D}$ promotes diversity as it would be small for data points which are similar to the points already selected in the previous sub-queries.

Now, we define an updated score $S$:

\begin{equation}
\label{eq:diversity_score}
    S(\textbf{x}_{u})=S_S(\textbf{x}_{u}) +  \lambda_{1} S_C(\textbf{x}_{u}) + \lambda_{2} S_{D}(\textbf{x}_{u})
\end{equation}
where $S_S(\textbf{x}_{u})$ and $S_C(\textbf{x}_{u})$ are the previously defined confusion score components. Using Equation~\ref{eq:diversity_score} we select another sub-query of $\frac{N}{K}$ samples, and the process repeats until we have $N$ samples. 

The process of selecting the query samples $\mathcal{D}_Q$ is described in Algorithm \ref{alg:pal} dubbed \emph{pretext-based active learning (PAL)}. While $g_{\phi'}$ is fine-tuned during a sub-query, all networks are trained from scratch using the cross entropy loss $\mathcal{L}$ after the oracle labels $\mathcal{D}_Q$.

\begin{algorithm}
\SetAlgoLined

\DecMargin{1em}

\KwResult{Set of additional samples to be labeled $\mathcal{D}_{Q}$}

\KwData{Labeled pool 
$\mathcal{D}_{L}:=\left\{\textbf{X}_{L}, Y_{L}\right\}$, unlabeled pool $\mathcal{D}_{U}:=\left\{\textbf{X}_{U}\right\}$, query size $N$}

\textbf{Set:} Num. epochs $E_Q$ and $E_S$, num. sub-queries $K$

\textbf{\# Training task and scoring networks}
 
\For{$t \in \{1, \hdots, E_{Q}\}$ }{

  \For{$\{\textbf{x}_{l},y_{l}\} \in \mathcal{D}_{L}$}{
  
      $\theta \leftarrow \theta - \eta \nabla_{\theta} \mathcal{L}(f_{\theta}(\textbf{x}_{l}),y_{l})$ \# Task network
      
      $\psi \leftarrow \psi - \eta \nabla_{\psi} \mathcal{L}(h_{\psi}(\textbf{x}_{l}),y_{l})$ \# $S_{C}$
      
      \For{$i \in \{0,1,2,3\}$}{
         $\phi \leftarrow \phi - \eta \nabla_{\phi} \mathcal{L}(g_{\phi}\left(\text{rot}_{90i}(\textbf{x}_{l})),i\right)$ \# $S_S$
      }
   }
}

\For{$\textbf{x}_{u} \in \mathcal{D}_{U}$} {
   Use $g,h$ to compute and save $S_{S}(\textbf{x}_{u})$, $S_{C}(\textbf{x}_{u})$
}

\textbf{\# Diversity-based sub-query sampling}

\textbf{Initialize:} $\mathcal{D}_{Q}=\emptyset; \phi'=\phi$

\For{$k \in \{1, \hdots, K\}$}{
   \For{$n \in \{1, \hdots, \frac{N}{K}\}$}{
      \eIf{$k == 1$}{
         \(\textbf{x}_{q} \leftarrow \arg\min \limits_{\textbf{x}_{u}\in \mathcal{D}_{U}} S_{S}(\textbf{x}_{u})+\lambda_{1}S_{C}(\textbf{x}_{u}) \)
      }{
         \(\textbf{x}_{q} \leftarrow \arg\min \limits_{\textbf{x}_{u}\in \mathcal{D}_{U}} S_{S}(\textbf{x}_{u})+\lambda_{1}S_{C}(\textbf{x}_{u})+\lambda_{2}S_{D}(\textbf{x}_{u}) \)
      }
      \(\mathcal{D}_{Q}\leftarrow \mathcal{D}_{Q} \cup\left\{\textbf{x}_{q}\right\}  \)
      
      \(\mathcal{D}_{U}\leftarrow \mathcal{D}_{U} - \left\{\textbf{x}_{q}\right\} \)
   }
   
   \For{$t \in \{1, \hdots, E_S\}$}{
      \For{$\textbf{x}_{q} \in \mathcal{D}_{Q}$}{
         \For{$i \in \{0,1,2,3\}$}{
            \(\phi' \leftarrow \phi' - \eta \nabla_{\phi'} \mathcal{L}(g_{\phi'}\left(\text{rot}_{90i}(\textbf{x}_{q})),i\right) \) 
         }
      }
   }
   
   \For{ $\textbf{x}_{u} \in \mathcal{D}_{U}$ }{
      Use $g_{\phi'}$ to compute and save $S_{D}(\textbf{x}_{u})$
   }
}
Get oracle to label $\mathcal{D}_{Q}$ and update $\mathcal{D}_{L}$
\caption{Pretext-based Active Learning (PAL)} 
\label{alg:pal} 
\end{algorithm}

\section{Experiments and Results}

In this section, we empirically show the effectiveness of the proposed \emph{pretext-based active learning (PAL)} scheme. We discuss the experimental setup, datasets used, techniques compared, and the implementation details.

\textbf{Datasets:} We performed experiments on four datasets: (1) SVHN \cite{svhn}, where classification task has to be performed for ten digit classes (house numbers) with color images of size $32 \times 32$ pixels from google street view images, (2) CIFAR-10 \cite{krizhevsky09learningmultiple}, where classification task has to be performed on ten classes in this widely-used computer vision benchmark that contains color images of size $32 \times 32$ pixels, (3) Caltech-101 \cite{caltech101}, where classification has to be performed on color images of size $300 \times 200$ pixels belonging to 101 different classes, with between only 40 to 800 images per class, and (4) Cityscapes \cite{Cordts2016Cityscapes} where semantic segmentation has to be performed on images of size $2048 \times 1024$, with each pixel needing to be classified into one of 19 classes. 

\textbf{Techniques compared:} We compared the performance of our approach with the following active learning  strategies. (1) \emph{Random sampling:} This is the simplest but nevertheless a strong baseline involving randomly picking samples to be labeled. (2) \emph{Entropy:} This is a classical method where the sample uncertainty is modeled as the entropy of its predicted class probabilities. (3) \emph{VAAL:} This technique uses a VAE to learn a feature space and then adversarially trains a discriminator on it~\cite{Sinha_2019_ICCV}. (4) \emph{DBAL:} This method uses Bayesian CNNs to estimate uncertainty (novelty) of unlabeled points~\cite{gal2017deep}. (5) \emph{Core-set:} This is a representation-based method for selecting the samples most different than the labeled samples and seeks to maximize the diversity of the samples to be picked for labeling~\cite{sener2018active}.
 
\textbf{Experimental setup:} Comparison between various active learning techniques was performed using a common experimental schema, in line with prior works~\cite{Sinha_2019_ICCV,sener2018active}. All techniques were used to iteratively expand the labeled dataset for training a common classifier architecture -- VGG16 \cite{vgg16} or a common semantic segmentation architecture- Deeplabv3 \cite{chen2017rethinking} with a MobileNetv2 \cite{sandler2019mobilenetv2} backbone  -- from scratch during each query round. The average accuracy of five random initializations were computed. The initial labeled pool of samples was shared by all techniques. For the image classification datasets, the initial labeled pool comprised 10\% of the whole dataset, and each query round added an additional 5\% of the samples selected by the individual active learning technique. For semantic segmentation, the initial labeled pool consisted of 5\% of the total dataset, each query round added an additional 1\% samples to the labeled dataset, and mean intersection over union (mIoU) was used as the performance metric.

For the scoring network of the proposed PAL approach, we used a ResNet-18 \cite{resnet18} architecture. Using validation, the relative importance hyperparameters in Equation~\ref{eq:diversity_score}) were selected from \{0.5 , 1.0\}. Learning rates were in the range [$10^{-1}, 10^{-4}$]. Optimizers were selected from \{ADAM, SGD\}. The hardware included an NVIDIA GeForce GTX 1080 GPU running CUDA 10.2 and cuDNN 7.6 using PyTorch.

\subsection{Performance with error-free labels}
\label{sec:main_res}

\begin{figure*}[h]
    \begin{minipage}{.48\textwidth}
    \centering
    \includegraphics[width=\textwidth]{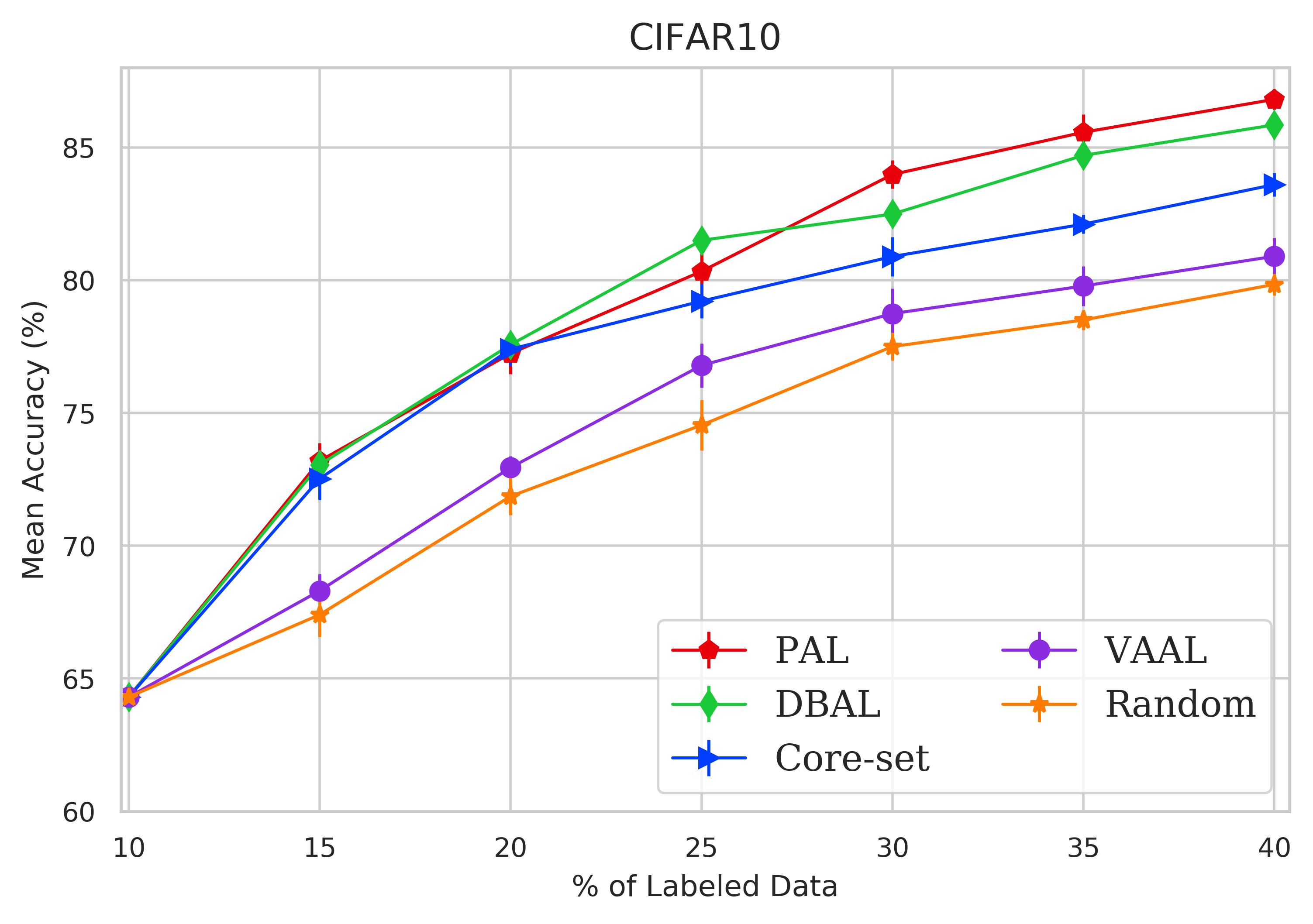}
    \label{fig:CIFAR-10 basic results}
    \end{minipage}
    \quad
    \begin{minipage}{.48\textwidth}
    \centering
    \includegraphics[width=\textwidth]{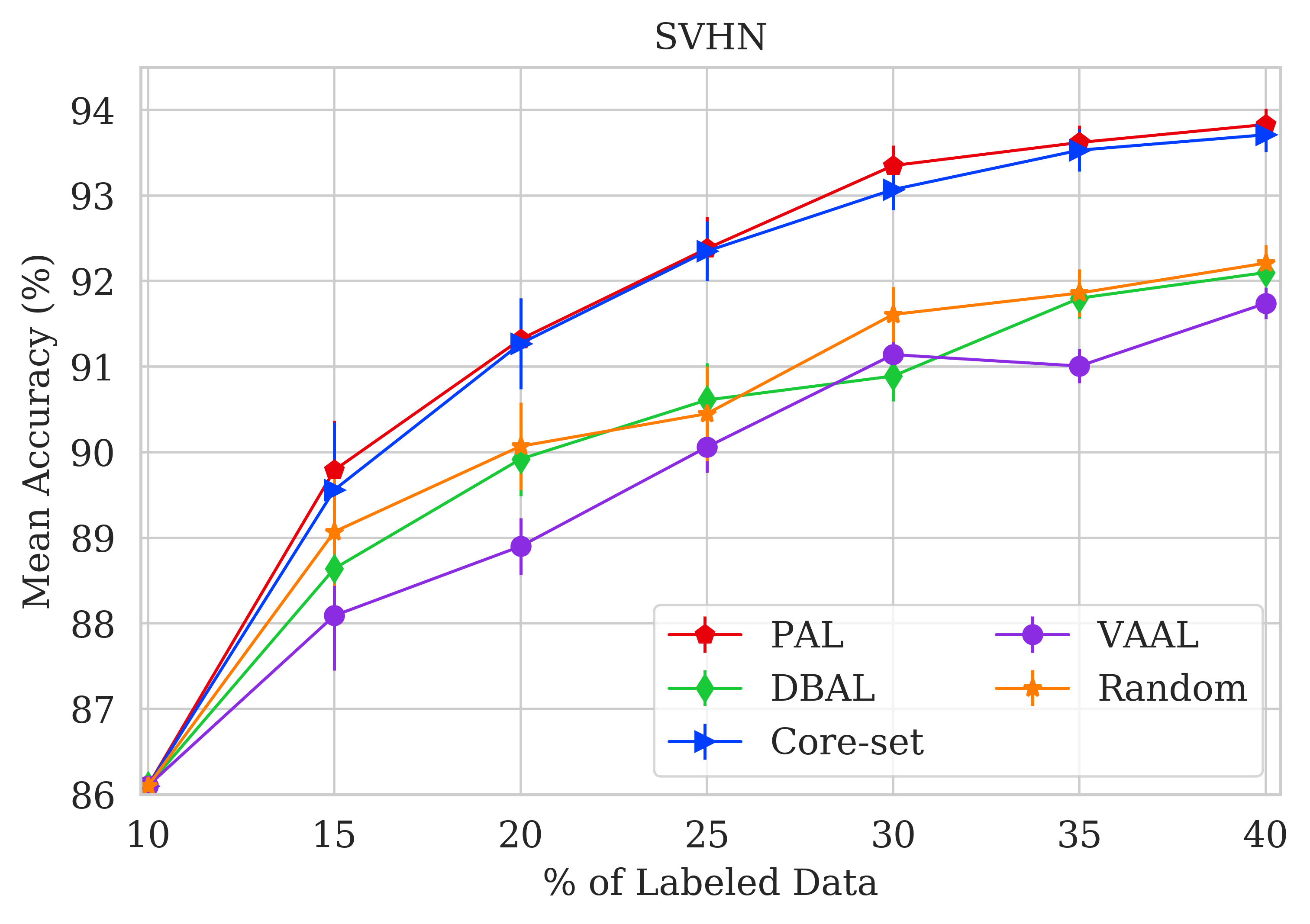}
    \label{fig:SVHN basic results}
    \end{minipage}
    \quad
    \begin{minipage}{.48\textwidth}
    \centering
    \includegraphics[width=\textwidth]{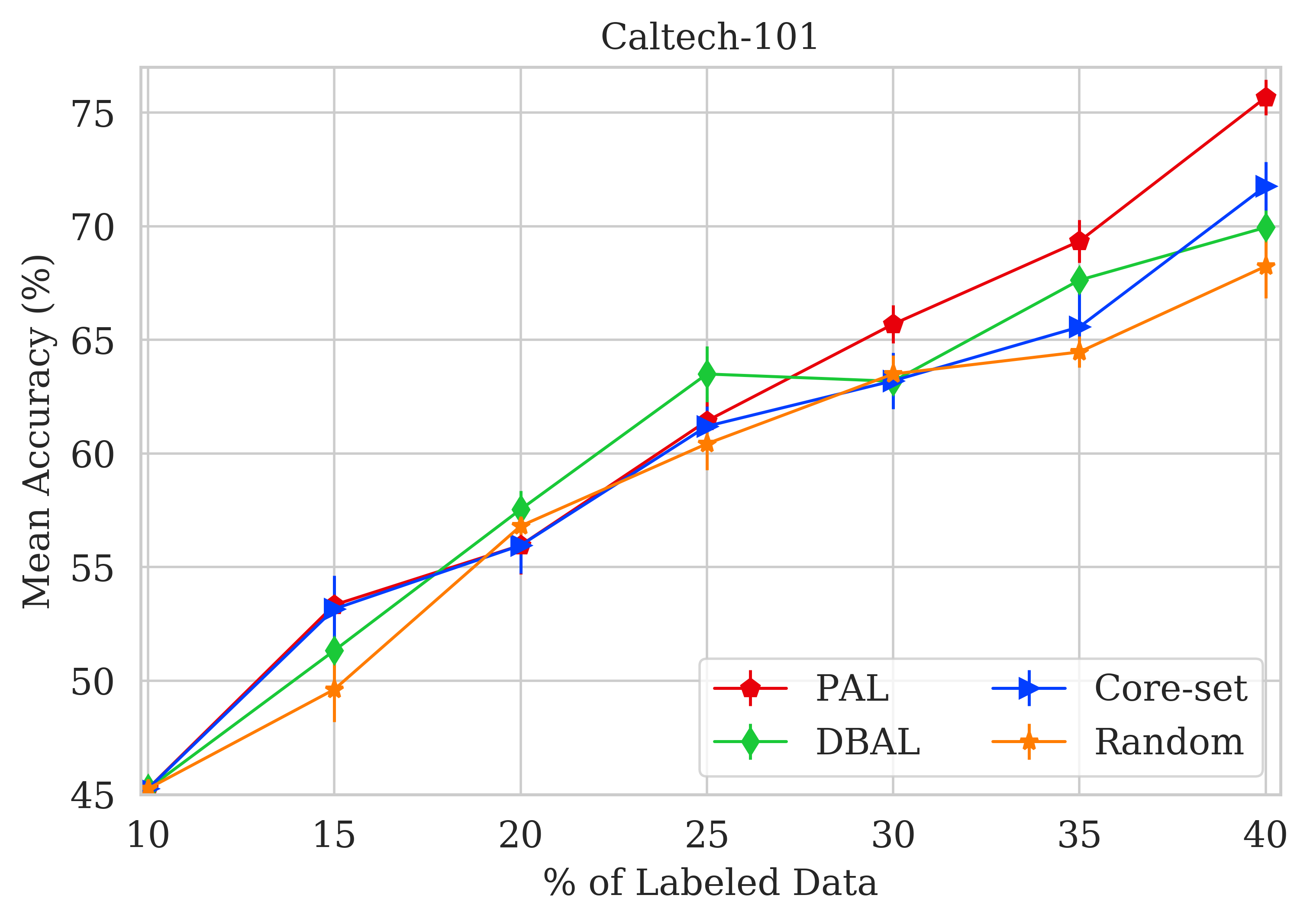}
    \label{fig:Caltech-101 basic results}
    \end{minipage}
    \begin{minipage}{.48\textwidth}
    \centering
    \includegraphics[width=1.07\textwidth,height=0.7\textwidth]{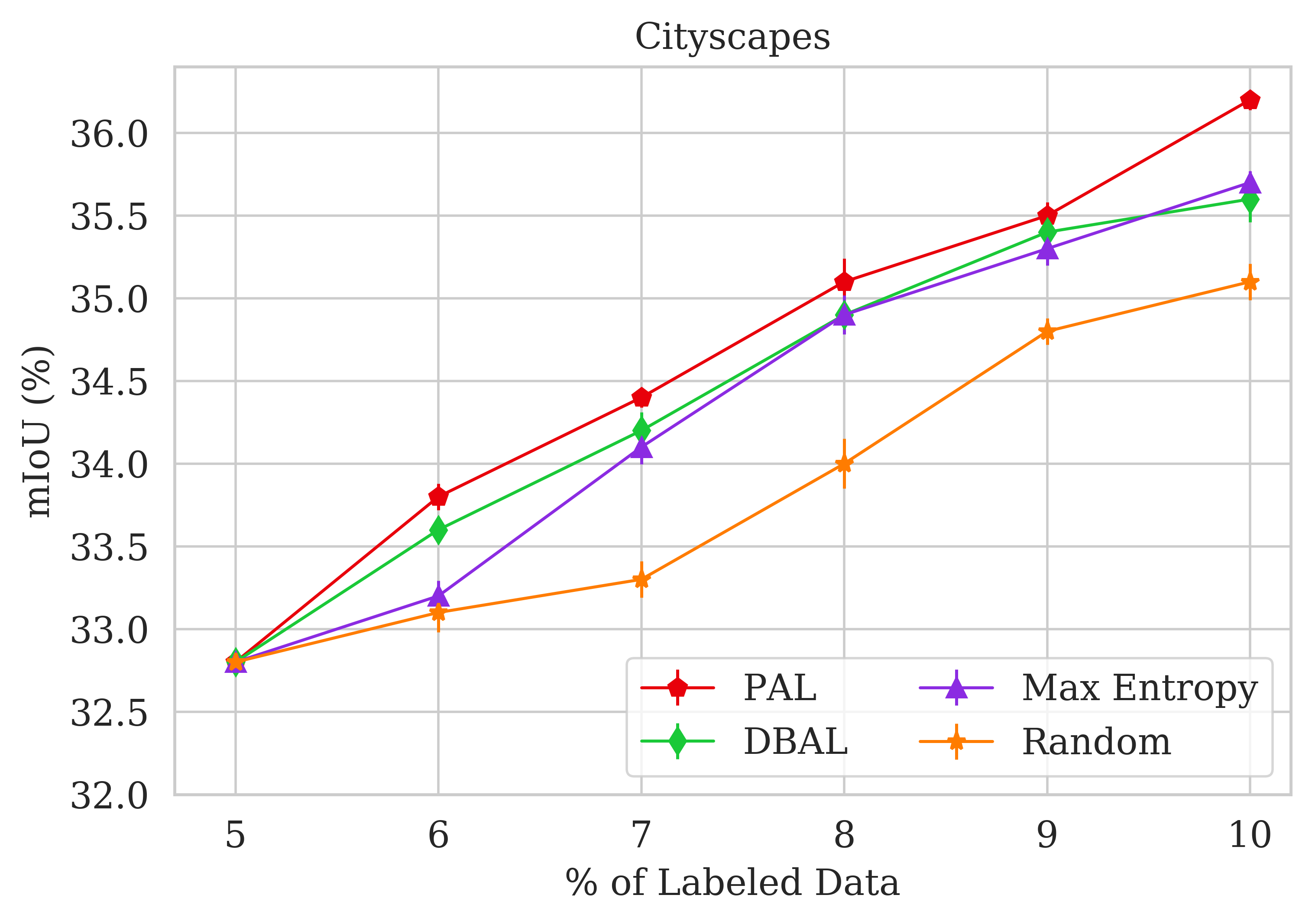}
    \label{fig:Cityscapes basic results}
    \end{minipage}
    \caption{Performance of  random sampling, entropy, VAAL \cite{Sinha_2019_ICCV}, DBAL \cite{gal2017deep}, and core-set \cite{sener2018active} compared with PAL (proposed) on CIFAR-10, SVHN, Caltech-101, and Cityscapes (segmentation). Markers show mean accuracy of five runs, and vertical bars show standard deviation (some are too small to be visible). \emph{*Note that VAAL takes prohibitively long to train due to the use of a VAE. Therefore, we did not train VAAL on Caltech-101 and CityScapes.}}
    \label{fig:main_res}
\end{figure*}

Figure~\ref{fig:main_res} compares the mean performance over five random initializations of different techniques for different fractions of the data labeled. Our PAL strategy outperformed random sampling by a wide margin and consistently seems to outperform VAAL \cite{Sinha_2019_ICCV}, DBAL \cite{gal2017deep}, and core-set \cite{sener2018active}. For instance, PAL requires only 20\% of labeled SVHN images to achieve performance equal to that achieved by VAAL and DBAL using 30\% labels, or a potential savings of 33\% labels. Additionally, PAL requires only about 2 hours per query round to train on a single 11GB GPU for SVHN, whereas more computationally expensive methods such as VAAL \cite{Sinha_2019_ICCV} take more than 24 hours for the same. Out of the techniques compared only core-set \cite{sener2018active} was faster than PAL, but its relative accuracy was quite variable across the datasets. Similar trends can be observed for semantic segmentation on CityScapes.

\subsection{Robustness to sample mislabeling}\label{sec:robust}

\begin{figure*}[h]
     \begin{minipage}{.48\textwidth}
    \centering
        \includegraphics[width=\textwidth]{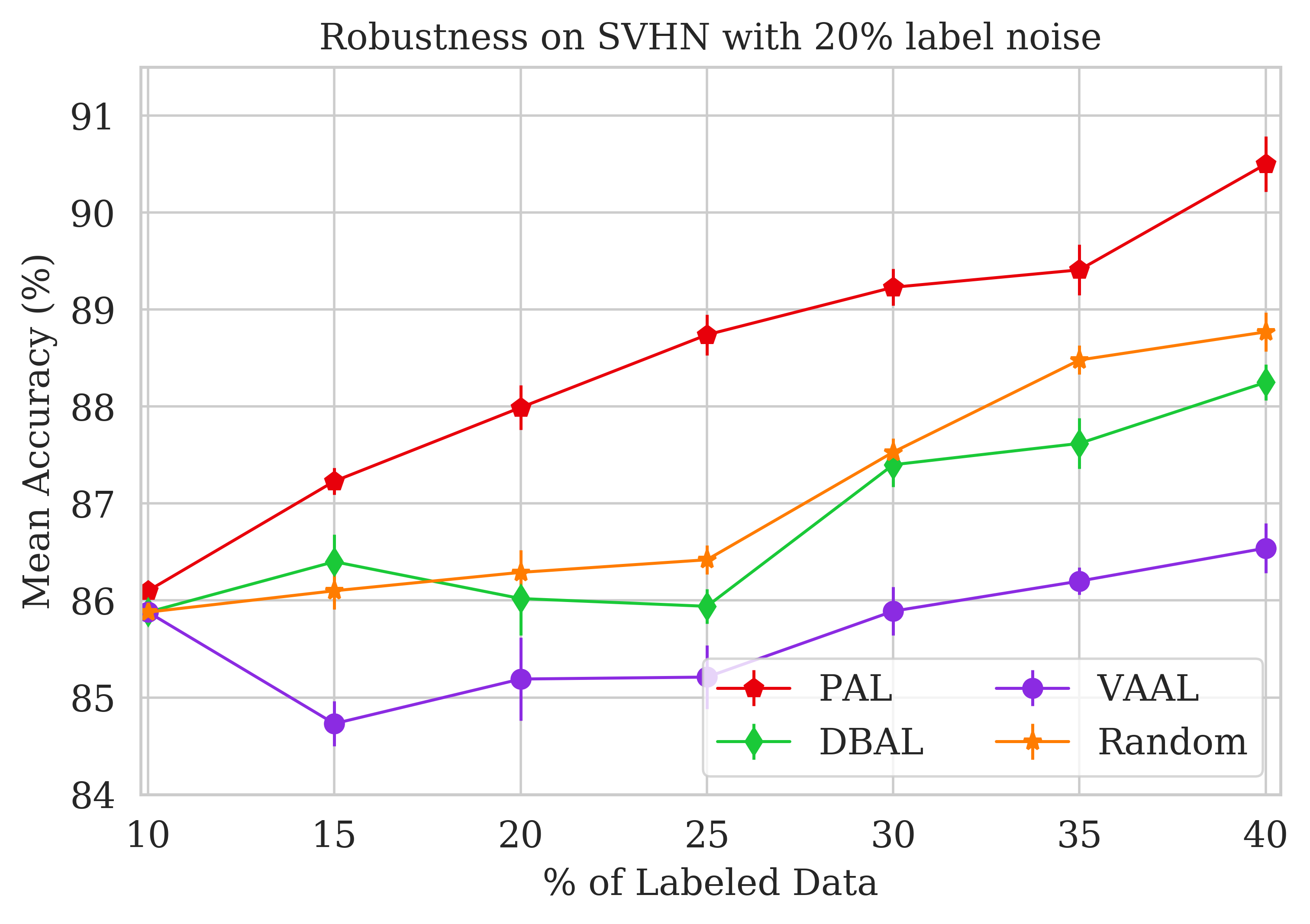}
        \end{minipage}
\quad
     \begin{minipage}{.48\textwidth}
        \centering
        \includegraphics[width=\textwidth]{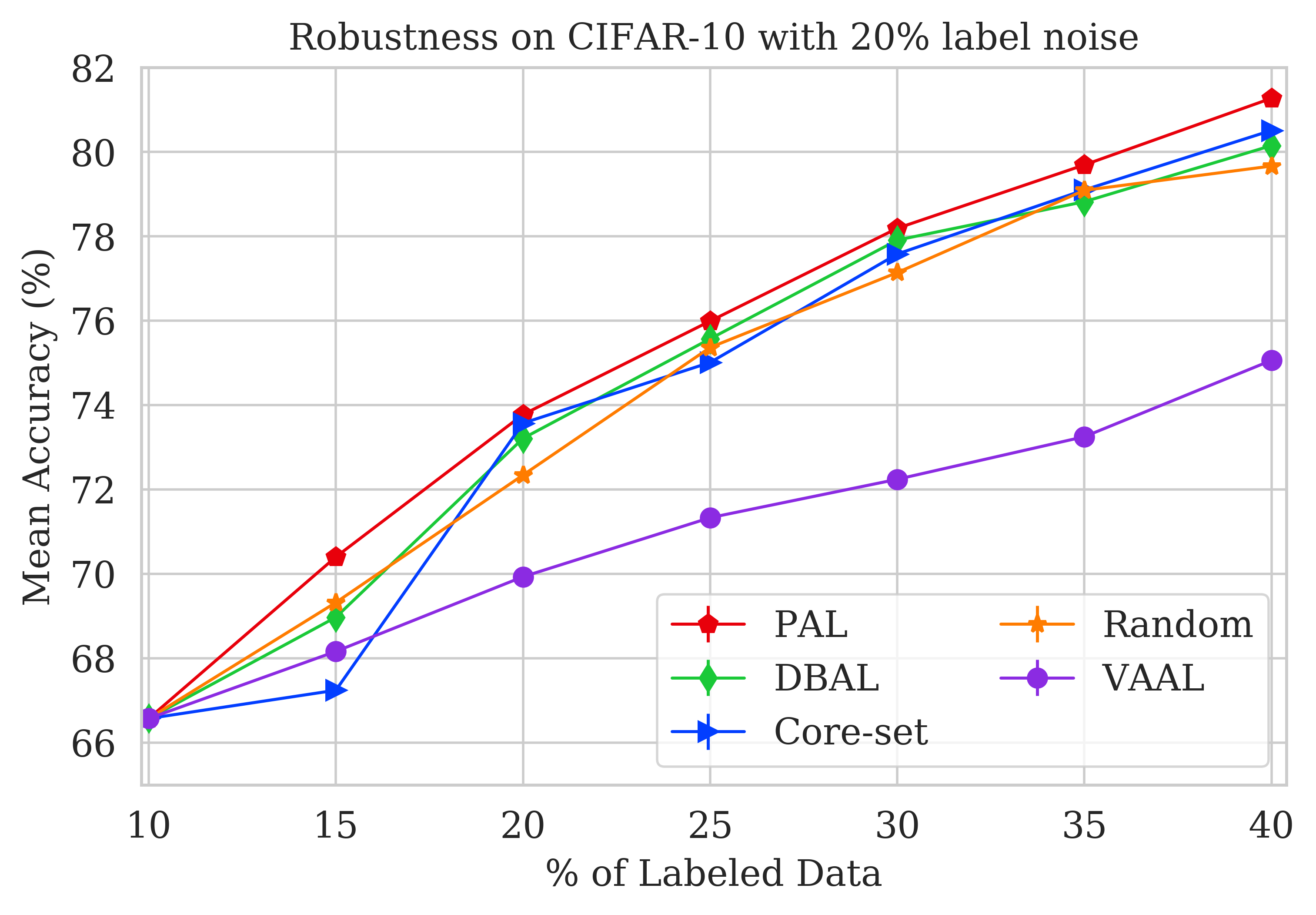}
    \end{minipage}
    
    \caption{Active learning techniques compared with 20\% label noise on SVHN (left) and CIFAR-10 (right).}
    \label{fig:labelingError}
\end{figure*}

We simulated labeling errors for classification by randomly assigning incorrect labels to a subset of the labeled pool and the queried set. We performed experiments on the SVHN and CIFAR-10 datasets, corrupting 20\% of the data labels. In Figure \ref{fig:labelingError}, we observe that our technique clearly fares better compared to the others tested. We attribute this robustness of PAL to the use of the pretext task in the scoring network. 

    

\subsection{Introducing new classes on-the-fly}\label{sec:bias_pool}

\begin{figure*}[h]
     \begin{minipage}{.48\textwidth}
        \centering
        \includegraphics[width=\textwidth]{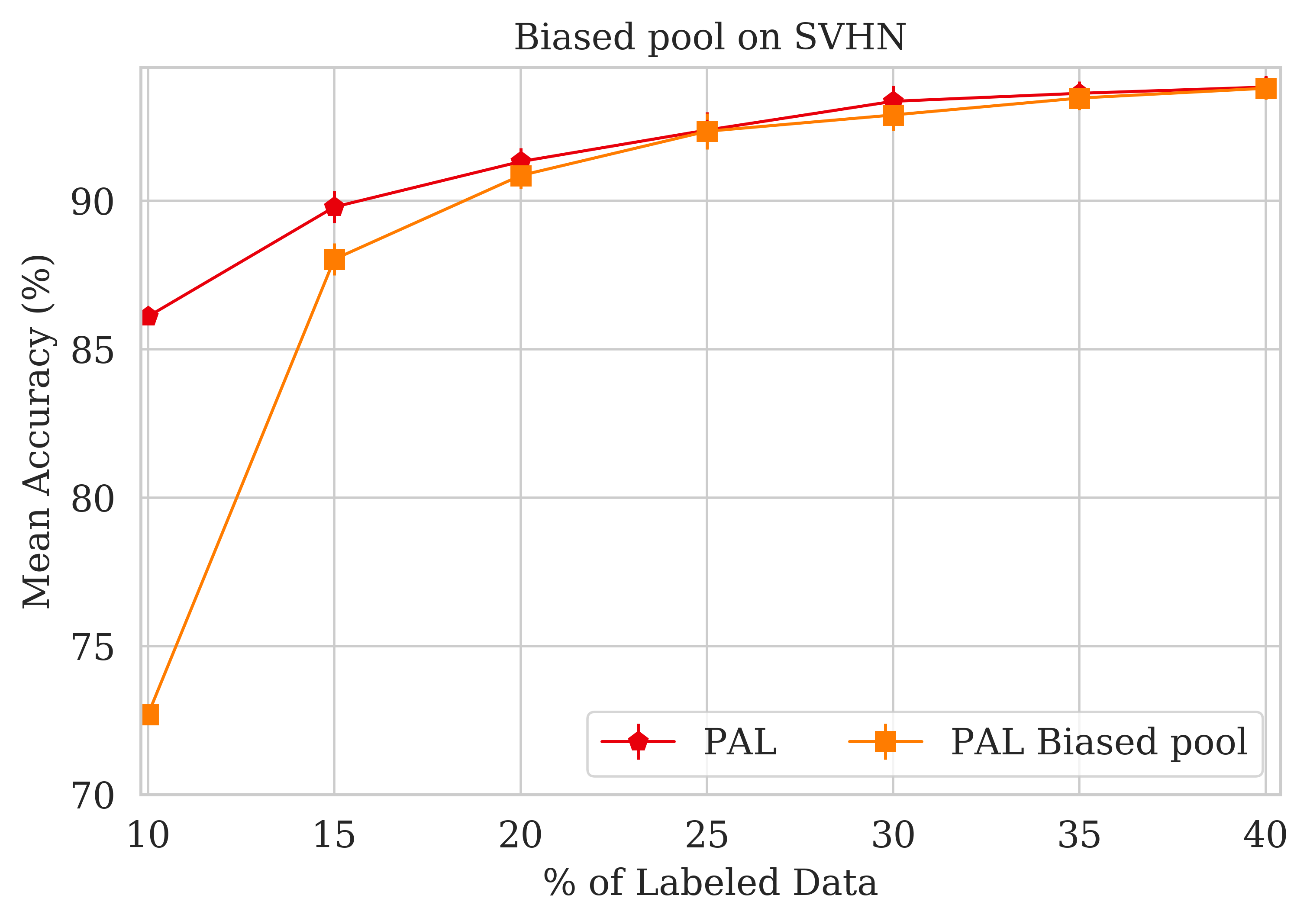}
    \end{minipage}
\quad
    \begin{minipage}{.48\textwidth}
        \centering
        \includegraphics[width=\textwidth]{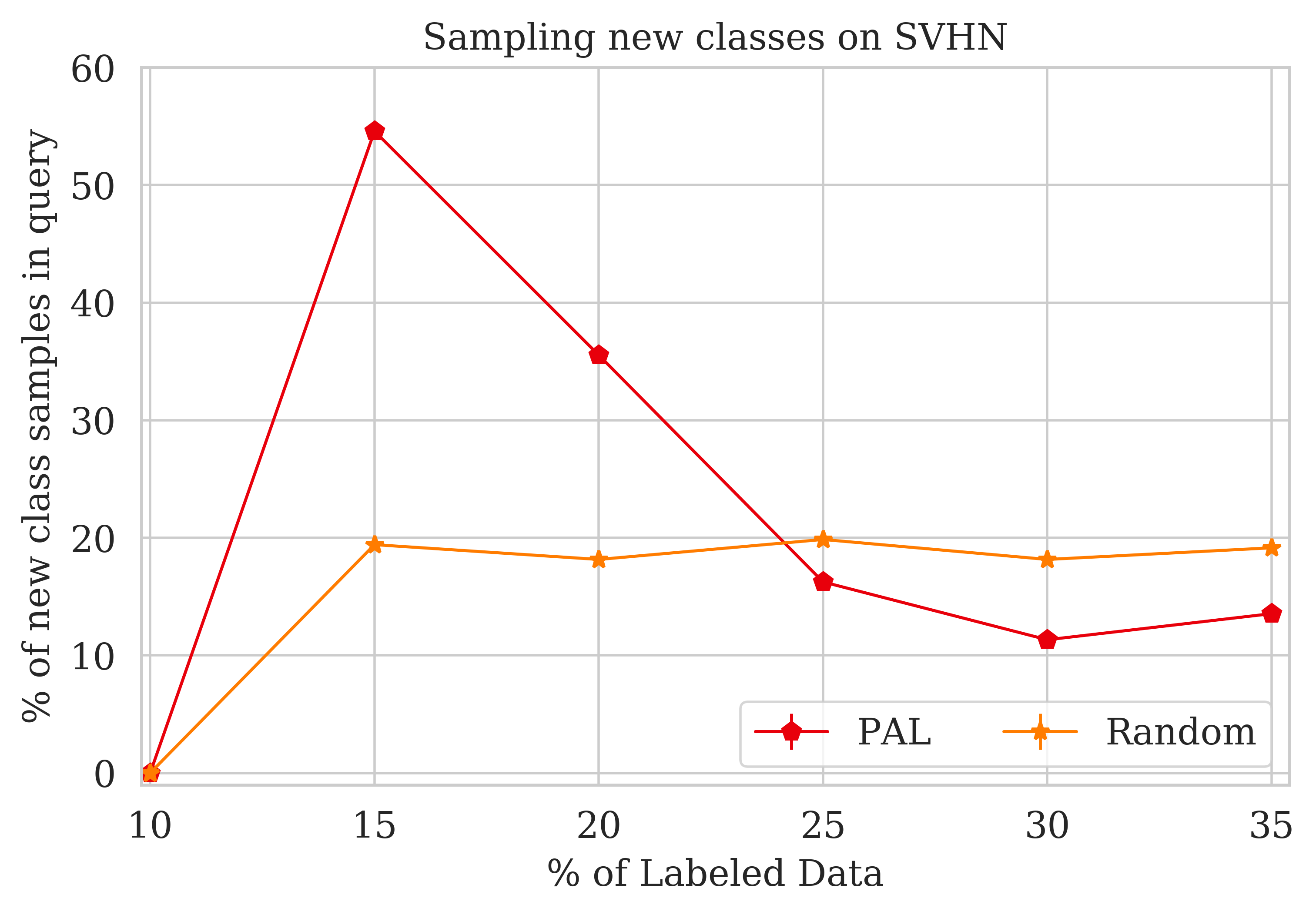}
    \end{minipage}
    
    \caption{PAL performance with biased initial pool of only eight out of ten classes: The accuracy of PAL trained with biased pool quickly catches up with that of the trained without the biased initial pool (left), because it temporarily oversamples the newly introduced two classes that it finds novel but the random sampling does not (right).}
    \label{fig:newClasses}
\end{figure*}

We performed experiments with a biased initial pool consisting of only eight out of the ten classes in the SVHN dataset. After the initial training, the algorithm was given access to unlabeled samples from all the ten classes to check its behavior. As seen in Figure~\ref{fig:newClasses}, PAL rapidly ramped up the performance when it was allowed to sample from the previously missing classes after the initial 10\% labels. In fact, it quickly caught up with its own strong performance on the unbiased initial pool case (i.e., the upper-left graph of Figure~\ref{fig:newClasses} is same as that of SVHN results in Figure~\ref{fig:main_res}). It temporarily over-sampled the previously missing classes, and the sampling returned to a balanced one after its performance caught up with its own version that was not initially deprived of the samples from the two missing classes. On the other hand, the representation of the two missing classes remains around 20\% for random sampling, once those classes are made available for queries, as expected. We observed similar trends for semantics segmentation on the Cityscapes dataset, where we started of with 17 out of the 19 classes and observed that PAL was able to select images which had upto seven times higher pixel area corresponding to missing classes compared to random. Please see Appendix A for the accuracy and the sampled pixel area for missing classes plots.

\subsection{Robustness to scoring network architecture}
\label{sec:Robust2Arch}
\begin{figure}[h]
    \centering
    \includegraphics[width=0.48\textwidth]{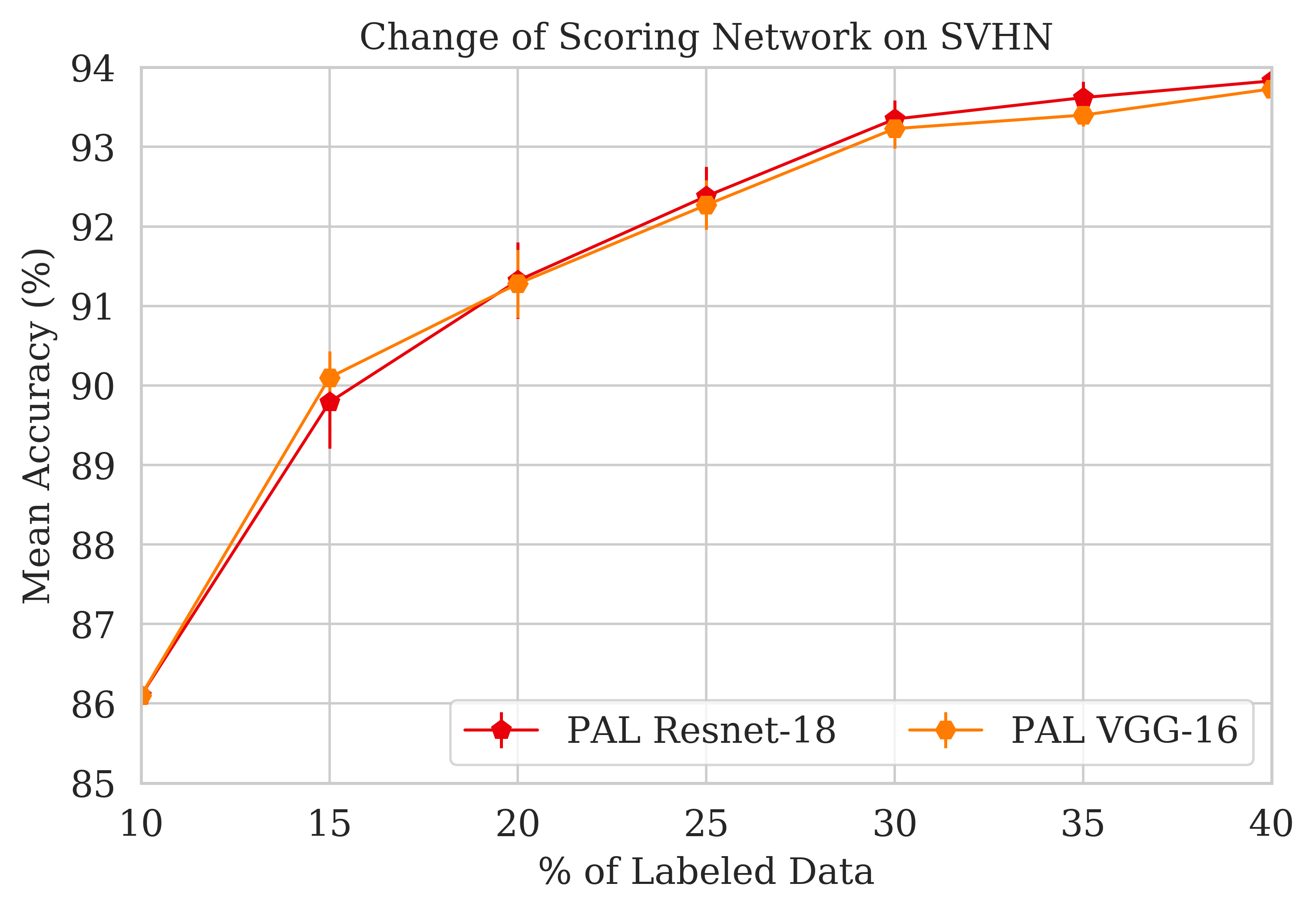}
    
   
    \caption{Performance of PAL with Resnet-18 and VGG-16 as backbones of the scoring network are not that much different from each other on the SVHN data.}
    \label{fig:SVHNmodchange}
\end{figure}

We now show that the unique features of PAL -- separate scoring network, self-supervised learning, multi-task scoring network, and diversity in sampling -- seem to be more responsible for its strong performance than the backbone architecture of the scoring network. We replaced ResNet-18 with VGG-16, and found no significant change in performance on the SVHN dataset, as shown in Figure \ref{fig:SVHNmodchange}.

\subsection{Importance of the score components} 
\label{sec:ablation}
\begin{figure}[h]
    \centering
    \includegraphics[width=0.48\textwidth]{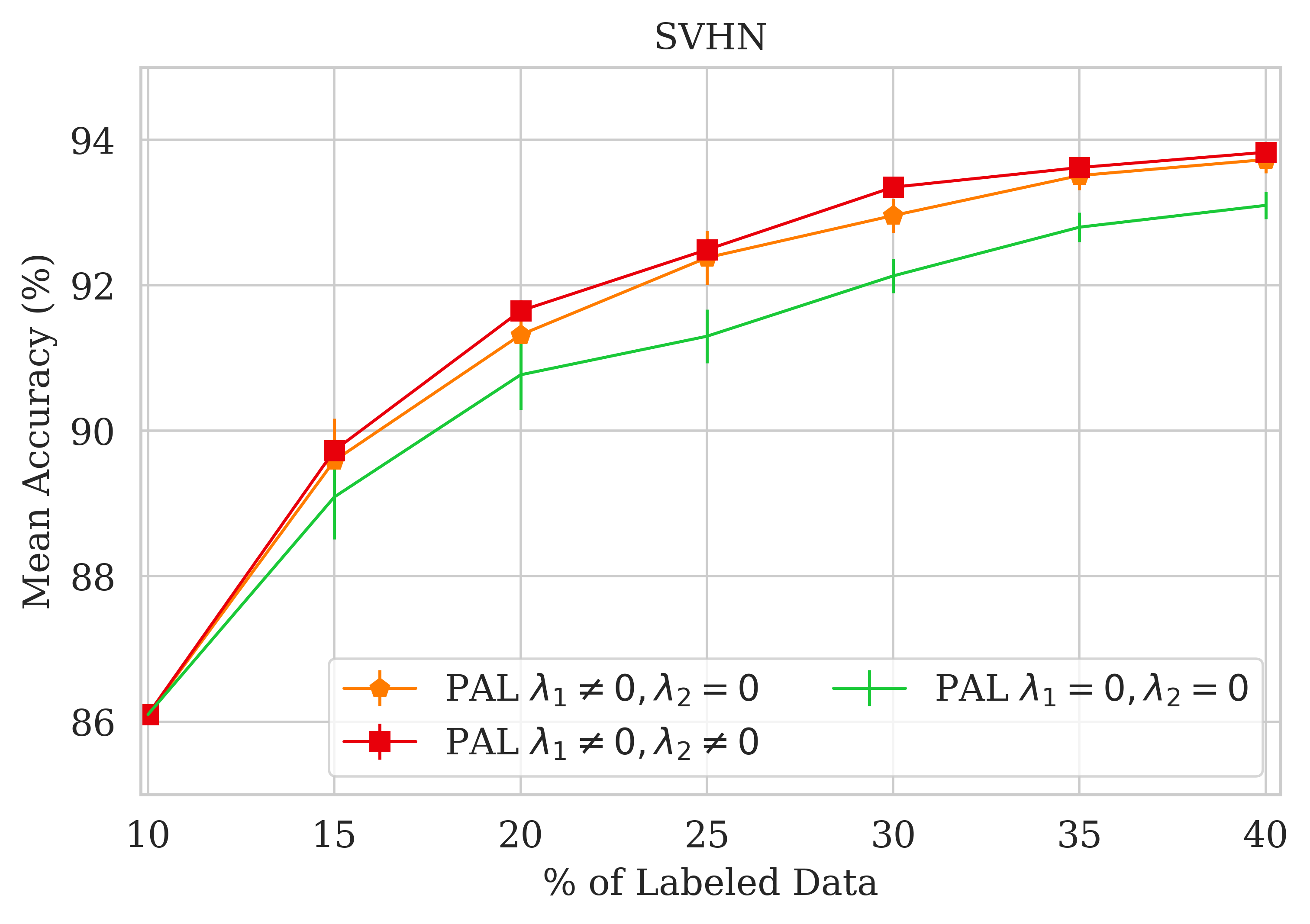}
    
   
    \caption{Contribution of PAL's components for scoring unlabeled samples compared: using only self-supervision ($\lambda_{1}=0,\lambda_{2}=0$), self-supervision and supervision ($\lambda_{1} > 0, \lambda_{2}=0 $), and self-supervision with supervision and sub-query-based diversity  ($\lambda_{1} > 0, \lambda_{2} > 0$).}
    \label{fig:SVHNabl1}
\end{figure}

We examined the effect of the different components of the proposed score in Equation~\ref{eq:diversity_score} used to formulate the query by performing an ablation study. We compared performance by dropping the diversity score $S_D$ ($\lambda_{2}=0, \lambda_{1} > 0$), dropping both the diversity score $S_D$ and the supervision score $S_C$ ($\lambda_{1}=0, \lambda_{2}=0$), and the original scenario with both diversity and supervision included ($\lambda_{1} >  0, \lambda_{2} > 0 $). We observed that using uncertainty estimates from both the pretext and classification tasks gave a much better performance. Adding the diversity score resulted in a further improvement in the performance. These results are shown in Figure~\ref{fig:SVHNabl1}, and they suggest that all the three components are important.

\begin{figure}[h]
    \centering
    \includegraphics[trim=0cm 0cm 0cm 0cm, clip, width=0.48\textwidth]{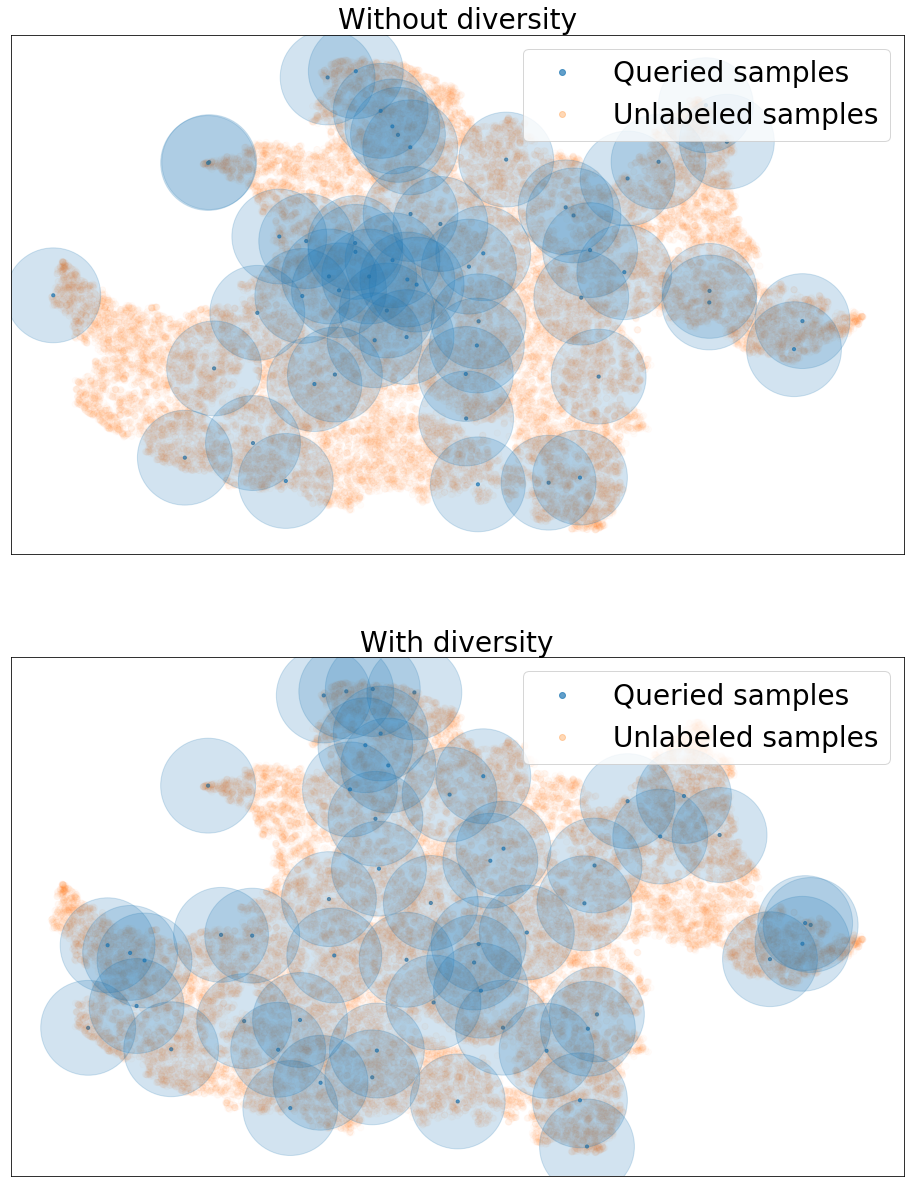}
    
   
    \caption{A comparison of the coverage (blue circles) of the unlabeled data (orange points) from CIFAR-10 using VGG-16 network and PAL-based query points (blue points) selected (a) without and (b) with diversity show more crowding in the former, and better coverage in the latter.}
    \label{fig:diversity}
\end{figure}

Additionally, we visually show that dividing the query into sub-queries indeed increases the diversity of the query. In Figure \ref{fig:diversity}, we can see two t-sne embedding plots~\cite{van2008visualizing} for CIFAR-10 dataset using a VGG-16 network trained on 10\% of data. The unlabeled samples are shown in with orange color. Selected query points without and with diversity are shown in the top and bottom sub-plots respectively with blue dots. A blue circle was included for each query point centered at its location to visualize its sphere of coverage. It seems clear that without diversity, there are gaps in coverage in some areas and crowding of query points in other areas. With diversity, the query points are more spread out, and provide better coverage of the unlabeled points.

\section{Conclusion and Discussion}
\label{sec:conclusion}

We proposed a new pool-based active learning method that is robust to partial mislabeling of the training data, while also giving competitive results for the correctly labeled data. It uses a separate sample scoring network to resolve conflicts in the task goal and uncertainty estimation. It relies on self-supervised learning to reduce the dependence on potentially mislabeled data. It also uses a multi-task setting to regularize the scoring network, and to compensate for the failures of self-supervised learning. The scoring network is trained on the labeled samples in order to model their distribution instead of that of the entire data. Further, it ensures diversity between the sub-queries of a query. 

There is a need to balance between the twin goals of assessing novelty and diversity to select samples for the queries. By relying on novelty alone, there is a danger of picking a lot of novel samples that do not reasonably cover all regions of the data distribution. Conversely, methods that rely on diversity, such as core-set~\cite{sener2018active}, can be hijacked by outliers in higher dimensions. While more research is needed to jointly pursue both goals, our method takes a sub-query approach to ensure that at least the samples in a sub-query are different from that of another sub-query.

Each of its component seems to contribute to its performance. It also takes a reasonable time to train, as it requires only one additional network trained in a regular supervised manner. This work presents early evidence that over-reliance on only one measure of uncertainty may not be judicious, and hybrid methods, where individual components compensate for each other, are likely to work better. A hybrid scoring method is able to break the reliance on labels, which may be noisy, for modeling the data distribution by going lower in the semantic hierarchy and tapping into the knowledge gained by self-supervision. A similar observation that self-supervised tasks add robustness in anomaly detection tasks has been reported previously~\cite{hendrycks2019using}. We empirically validated our hypotheses by showing strong performance of PAL on a variety of datasets. Furthermore, we also showed that PAL performs well even when the initial labeled data pool has no samples from a few of the classes.

\bibliographystyle{ieee_fullname}
\bibliography{local}
\clearpage
\appendix







\onecolumn
\section{Appendix A}
In Section 4.3 we evaluated PAL in the setting when new classes are introduced on-the-fly during the active learning based sampling. We started of with a biased initial pool where some of the classes are removed in the initial labeled data pool. Here we show the results in the same setting on a segmentation task on Cityscapes dataset. Out of the 19 classes in Cityscapes dataset, we removed the annotations of the bus and the train classes from the initial labeled data pool. For clarity, all further query rounds have access to all the class annotations. 
\par
\begin{figure*}[h]
    \begin{minipage}{.48\textwidth}
    \centering
    \includegraphics[width=\textwidth]{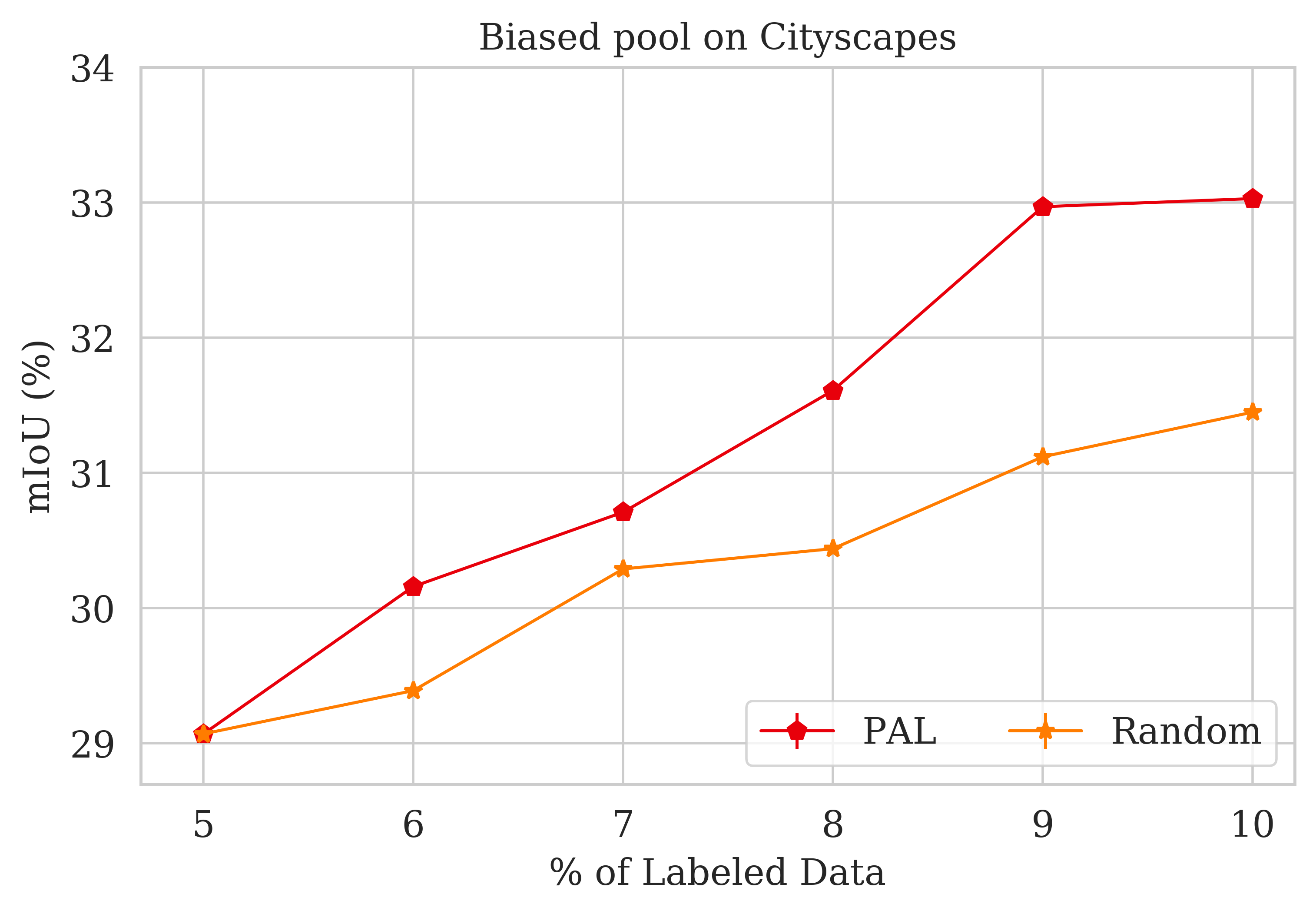}
    \end{minipage}
    \quad
    \begin{minipage}{.48\textwidth}
    \centering
    \includegraphics[width=\textwidth]{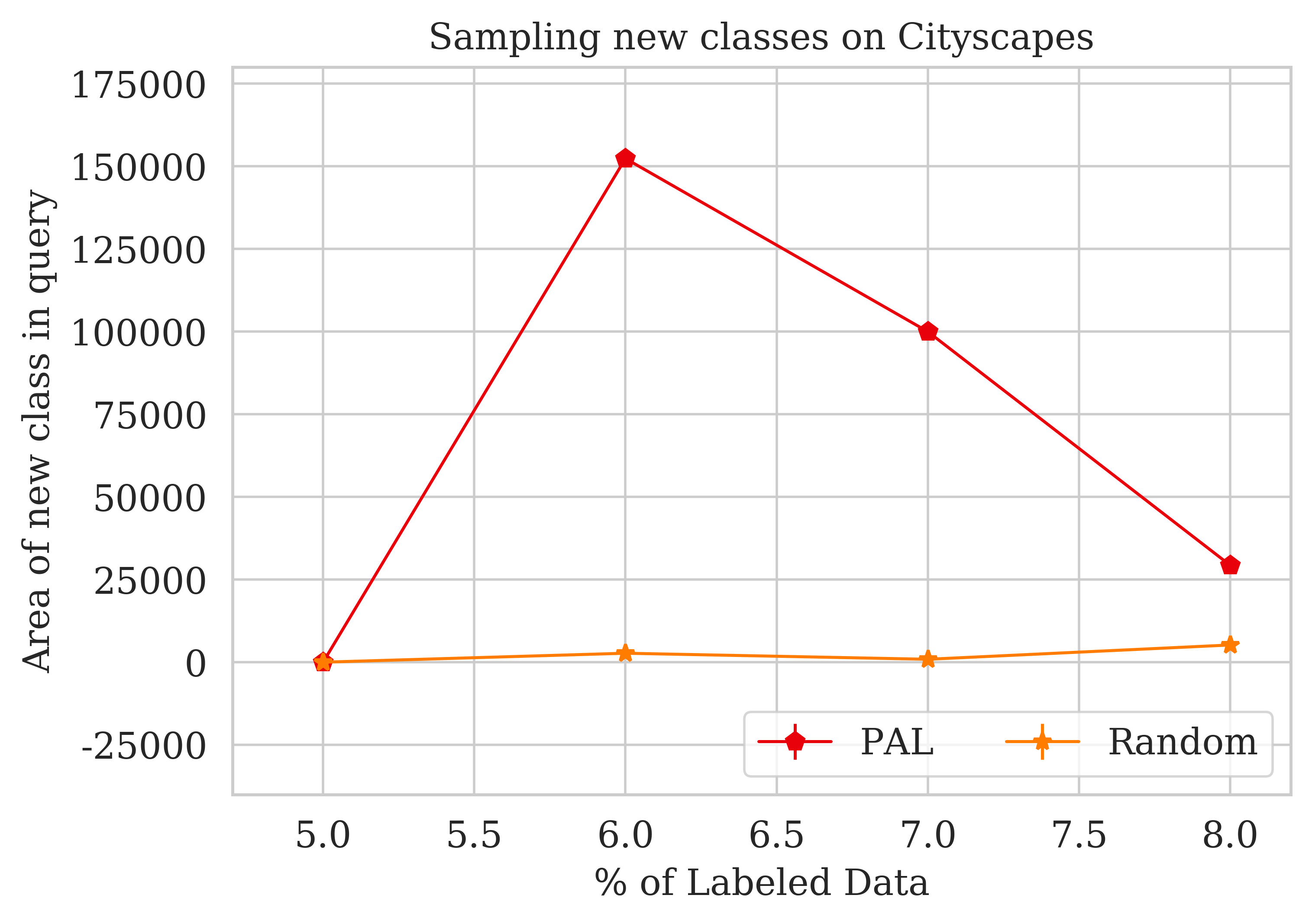}
    \end{minipage}
    \quad
    \caption{PAL performance with biased initial pool of only seventeen out of nineteen classes: The mIoU of PAL trained with biased initial pool improves quickly as compared to that of random sampling (left), because it is able to sample more images with higher area of missing annotations compared to random sampling (right)}
    \label{fig:supp_res}
\end{figure*}

 Figure \ref{fig:supp_res} (left) compares the mIoU of PAL with random sampling in this setting. PAL is able to improve its mIoU quickly because it is able to sample more images with higher area of missing annotations compared to random sampling as shown in Figure \ref{fig:supp_res} (right).

\section{Appendix B}
We share the hyperparameters used for training the task and the scoring models for our different experiments in Table \ref{tab:datahyper}. All hyperparameters were obtained through a grid search.
The hyperparameters $\lambda_{1}$ and $\lambda_{2}$ of Equation 4 in Section 3.3 were selected from \{0.5 , 1.0\}. Learning rates $\alpha_{T}$ for the task model and $\alpha_{S}$ for the scoring model were selected from the range [$10^{-1}, 10^{-4}$]. Optimizers were selected from \{Adam, SGD\}.
\\

\begin{table*}[h]
    \begin{tabular}{|c|c|c|c|c|c|c|c|}
    \hline
         \textbf{Dataset} & \textbf{$\alpha_{T}$} & \textbf{$\alpha_{S}$} & \textbf{\begin{tabular}[c]{@{}l@{}}task \& \\ scoring \\ model\\ epochs\end{tabular}} & \textbf{\begin{tabular}[c]{@{}l@{}}batch\\ size\end{tabular}} & \textbf{$\lambda_{1}$} & \textbf{$\lambda_{2}$} & \textbf{optimizer}    \\ \hline
         CIFAR-10 & 0.01 & 0.01 & 100 & 64 & 1 & 1 & SGD \\ \hline
         SVHN & 0.01 & 0.01  & 100 & 64 & 1 & 1 & SGD \\ \hline
         Cityscapes & 0.01 & 0.01 & 50 & 8 & 0.5 & 0 & SGD \\ \hline
         Caltech-101 & 0.01 & 0.01  & 100 & 32 & 1 & 1 & SGD \\ \hline
    \end{tabular}

    \caption{Parameters for experiments on various datasets}
    \label{tab:datahyper}
\end{table*}

\section{Appendix C}

\textbf{Proposition 1: } \textit{Negative of KL-divergence of a class PMF from a uniform distribution can overshadow the confusion score from $S_S$, but entropy cannot.}
\\
\textbf{Proof: }Consider a binary classification problem for analysis, with $p$ as the predicted probability score by the task network for the correct class. When the unlabeled sample is almost correctly classified with $p \to 1$, we get the following for the hybrid confusion score:
\[
\begin{split}
 S(\textbf{x}_{u}) & =S_S(\textbf{x}_{u}) +  \lambda S_C(\textbf{x}_{u}) \label{eq:score1} \\
\lim_{p \to 1} S & = \lim_{p \to 1} \left(S_S - \frac{\lambda}{2}\  \log \left(\frac{1}{2p} \right) -\frac{\lambda}{2}\  \log \left(\frac{1}{2(1-p)}\right)\right)
\\
& =S_{S} -\frac{\lambda}{2}  \log\left(\frac{1}{2}\right) - \frac{\lambda}{2}  \lim_{p \to 1}\log \left(\frac{1}{2(1-p)} \right)
\\
& = -\infty.
\end{split}
\]
On the other hand, if $S_C$ is replaced by the entropy of the PMF $h_{\psi}$, then the hybrid score $S_E$ would be finite because: 
\[
\begin{split}
    \lim_{p \to 1} S_E & = \lim_{p \to 1} \left(S_S- \lambda p \log(p) - \lambda (1-p) \log(1-p) \right)\\
    & =S_S - 0 - \lambda \lim_{p \to 1}  (1-p) \log(1-p)\\
    & =S_{S} - \lambda \lim_{p \to 1} \frac{\log(1-p)}{\frac{1}{(1-p)}} = S_S,
\end{split}
\]
using L'H\^{o}pital's rule to equate the second term to $0$. \hfill $\Box$

An added advantage of using a multi-task setting for the scoring network is getting better ordinal estimates of a true latent score due to an ensemble-like effect, as long as the correlations between the two components of the score and their correlation with the underlying score are positive. This can follow from the following proposition:

\textbf{Proposition 2: } \textit{There exists a trade-off parameter that maximizes the correlation between the true underlying score and the hybrid score, which is greater than or equal to the correlation of the true score with either of the components, as long as all correlations between the scores are positive.}
\\
\textbf{Proof: }Note that the requirement of a positive correlation is only a weak one for any reasonably trained networks $g_{\phi}$ and $h_{\psi}$, as we empirically show in  Table 1 in the main paper. Now, without loss of generality, let us assume that some monotonic transformations of the true underlying score, the self-supervision score, and the classification score give standardized random variables $u$, $v$, and $w$ respectively, such that their means $\mu_{u}=\mu_{v}=\mu_{w}=0$, and their variances $\sigma^2_{u}=\sigma^2_{v}=\sigma^2_{w}=1$. Further, we assume that the covariances $\sigma_{uv}$, $\sigma_{uw}$, and  $\sigma_{vw}$ are positive. Let an analog of the hybrid score $s$ be a positive combination of the two given by $s=\alpha v + \sqrt{1-\alpha^2} w$, where $\alpha \in [0,1]$ has a monotonic relation with the $\lambda \geq 0$ in the jybrod score , and the variance $\sigma^2_{s}=1$. Then, the correlation between $u$ and $s$, which is the same as the cosine between them, is $\mathbf{E}\left[ u.s \right] = \alpha \sigma_{uv} + \sqrt{1-\alpha^2} \sigma_{uw} $. If we maximize this correlation by setting its derivative with respect to $\alpha$ to zero, we get:

\begin{align*}
 & \frac{d \mathbf{E}\left[ u.s \right]}{d \alpha} = 0 \\
\implies   & \frac{d}{d \alpha} \left( \sigma_{uv} \alpha + \sigma_{uw} \sqrt{1-\alpha^2} \right) = 0 \\
\implies  & \sigma_{uv} + \frac{-\alpha}{\sqrt{1-\alpha^2}} \sigma_{uw} = 0\\
\implies  & \sigma_{uv}^{2} (1-\alpha^2) = \sigma_{uw}^{2} \alpha^2 \\
\implies  & \alpha = \pm \frac{\sigma_{uv}}{\sqrt{\sigma_{uv}^{2}+\sigma_{uw}^{2}}}
\end{align*}

Clearly, a maxima for $\mathbf{E}\left[ u.s \right]$ exists, because its second derivative is negative for $\alpha^{*} = + \frac{\sigma_{uv}}{\sqrt{\sigma_{uv}^{2}+\sigma_{uw}^{2}}}$ when the covariances are positive, and $\alpha^{*} \in (0,1)$. \hfill $\Box$

\end{document}